\documentclass[10pt,twocolumn,letterpaper]{article}
\usepackage[accsupp]{axessibility}  
\usepackage{iccv}

\usepackage{epsfig}
\usepackage{graphicx}
\usepackage{amsmath}
\usepackage{amssymb, eucal, cite}

\usepackage{multirow}
\usepackage{tabularx}
\usepackage{arydshln}
\usepackage{pifont}
\usepackage{setspace}

\usepackage{graphicx}
\usepackage[lofdepth,lotdepth]{subfig}

\usepackage{booktabs}
\usepackage[accsupp]{axessibility}

\usepackage{makecell}

\usepackage[lined,vlined,ruled,commentsnumbered]{algorithm2e}   

\usepackage{upgreek}

\usepackage{url}

\usepackage{xcolor}
\usepackage[]{hyperref}
\hypersetup{
    colorlinks=true,
    urlcolor=blue,
    linkcolor=blue,
    citecolor=blue,
    linkbordercolor=blue
}

\usepackage[margin=4pt,font=footnotesize,labelfont=bf,labelsep=endash,tableposition=top]{caption}

\iccvfinalcopy 

\def\iccvPaperID{6510}

\ificcvfinal\fi

\begin{document}

\title{Meta Navigator:\\ Search for a Good Adaptation Policy for Few-shot Learning}

\author{
Chi Zhang$^{1}$
\quad
Henghui Ding$^{1,2}$
\quad
Guosheng Lin$^1$\thanks{Corresponding author: G. Lin (e-mail: {\tt gslin@ntu.edu.sg})}
\quad
Ruibo Li$^1$
\quad
Changhu Wang$^2$
\quad
Chunhua Shen$^3$
\\[0.2cm]
$^1$ Nanyang Technological University \qquad $^2$ ByteDance\qquad $^3$ Monash University

\\
{\tt\small chi007@e.ntu.edu.sg; 
gslin@ntu.edu.sg}
}
\maketitle

\maketitle

\ificcvfinal\thispagestyle{empty}\fi

\def\red{\textcolor{red}}

\begin{abstract}

Few-shot learning aims to adapt knowledge learned from previous tasks to novel tasks with only a limited amount of labeled data. Research literature on few-shot learning exhibits great diversity, while different algorithms often excel at different few-shot learning scenarios. It is therefore tricky to decide which learning strategies to use under different task conditions. Inspired by the recent success in Automated Machine Learning literature (AutoML), in this paper, we present Meta Navigator, a framework that attempts to solve the aforementioned limitation in few-shot learning by seeking a higher-level strategy and proffer to automate the selection from various few-shot learning designs. The goal of our work is to search for good parameter adaptation policies that are applied to different stages in the network for few-shot classification. We present a search space that covers many popular few-shot learning algorithms in the literature, and develop a differentiable searching and decoding algorithm based on meta-learning that supports gradient-based optimization. We demonstrate the effectiveness of our searching-based method on multiple benchmark datasets. Extensive experiments show that our approach significantly outperforms baselines and demonstrates performance advantages over many state-of-the-art methods.

\end{abstract}

\section{Introduction}
Convolutional Neural Networks (CNNs) have become indispensable in a variety of computer vision tasks~\cite{liu2020weakly,zhang2021cyclesegnet,sun2020conditional,zhang2018efficient,sun2020open,sun2021m2iosr,Zhang_2021_CVPR}.
A crucial reason is that the knowledge learned by CNNs can be transferred across different vision tasks in the form of hierarchical feature representations. Nevertheless, a sufficiently large amount of annotated data is still necessary to achieve good generalization accuracy due to CNNs’ data-hungry properties, which inevitably hinders the application of CNNs in real-world scenarios.

\begin{figure}[t]
	\centering
	\includegraphics[width=1\linewidth]{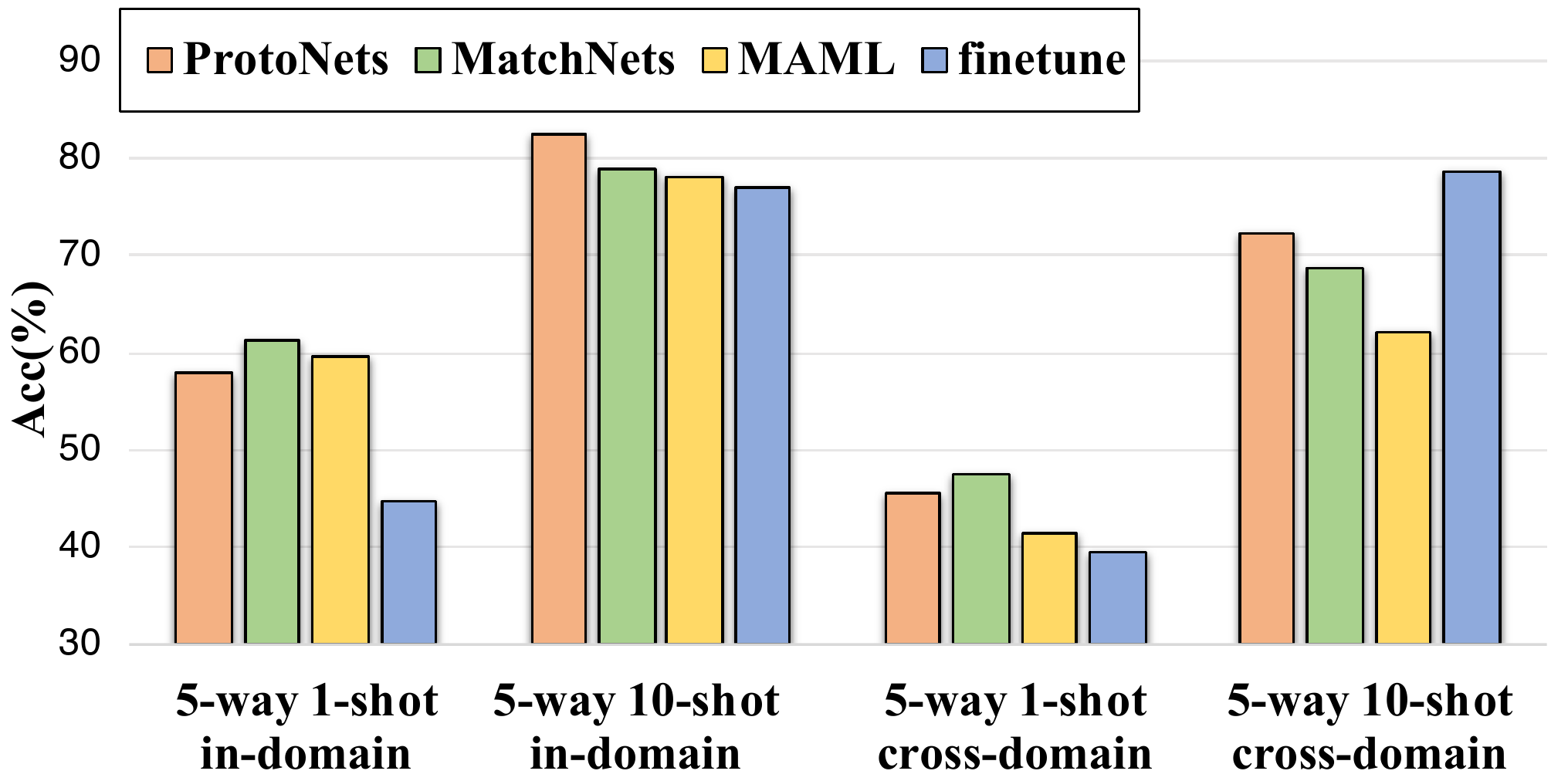}
	\caption{Comparison of some popular few-shot learning algorithms under various few-shot learning task settings. All models are based on the same network architecture, with weights pre-trained on \emph{mini}ImageNet dataset. Cross-domain experiments are evaluated on 
	the 
	CUB dataset.
	Existing few-shot learning algorithms are highly task-specific and are outperformed by a simple fine-tune baseline when the domain difference is large and more support data are available.}
	\label{fig:teaser}

\end{figure}

Few-shot learning is proposed as a promising direction to alleviate the need for exhaustively labeled data by exploring an extreme case where only a few labeled data is available to undertake a novel task based on prior knowledge learned on previous tasks. A typical application scenario is few-shot image classification \cite{VinyalsBLKW16,SnellSZ17,FinnAL17}.
Literature on few-shot learning exhibits great diversity, while different algorithms often excel at different few-shot learning scenarios.
Fig.~\ref{fig:teaser} compares some popular few-shot learning algorithms on different few-shot learning tasks.
Here we consider three test cases, including 1) the extreme low-shot case, \ie, 1-shot; 2) a medium-shot case, where the size of support set is relatively larger than the common benchmarks, \eg, 10-shot; 3) a cross-domain case where the training and testing tasks are sampled from different domains.
As is shown, existing few-shot learning algorithms are highly task-specific and no single algorithm can show superiority over others across all tasks.
In particular, when the domain difference is large, the compared few-shot learning algorithms can not sufficiently utilize the increasing number of support data to accommodate 
the domain difference, while the simple finetuning strategy can beat all other few-shot learning methods, although it is significantly outperformed by others in the 1-shot case due to over-fitting.
Therefore, it is almost impossible to find one single optimal few-shot learner that works well for all tasks.
This makes many few-shot learning algorithms difficult to be applied as a general tool to solve the data scarcity issue in machine learning, even though they can perform very well on some specific benchmarks.

In recent years, 
there is a surging interest 
in
automating the design of machine learning algorithms (AutoML), instead of
relying 
too much on
heuristic manual designs.
In particular, the idea of AutoML has been successfully applied to Neural Architecture Search (NAS)~\cite{liu2018darts,Zela2020Understanding,wang2021rethinking}, where the model learns to identify high-performance architectures by exploring a large candidate architecture space.
With the same intuition, in this work, we attempt to solve the aforementioned limitation in few-shot learning by seeking a higher-level strategy and take initiatives to automate the selection of few-shot learning designs.
The goal of our work is to search for 
good parameter adaptation policies that are applied to different stages in the network for few-shot learning. 
The search space in our network include two parts: the policy to adapt convolutional layers in different stages of the backbone and the policy to obtain class prototypes in the classifier, which together construct a hierarchical policy search space.
At each network stage, various candidates policies are available for adapting the parameters, and the whole search space covers many popular meta-learning algorithms in the literature, such as Prototypical Networks~\cite{SnellSZ17}, Matching Networks~\cite{VinyalsBLKW16}, baseline++~\cite{chen19closerfewshot}, MAML~\cite{FinnAL17}, \etc.

In order to search from a pool of discrete adaptation policies, we develop a differentiable searching algorithm based on meta-learning that allows 
efficient 
gradient-based optimization.
Inspired by the differentiable designs in NAS literature~\cite{liu2018darts}, our searching system is built upon a continuous relaxation of the discrete meta-learning policy, where each candidate policy is associated with a learnable policy selection 
indicator.
However, as each adaptation policy is an optimization process rather than a differentiable operation, directly porting the formulation in DARTS~\cite{liu2018darts} would not suffice. To tackle this issue, we further associate each policy with a group of policy-specific model parameters.
Then, the decision of choosing the optimal policy becomes jointly learning the policy selection indicators as well as the policy parameters.
The searching is conducted via a bi-level optimization paradigm based on meta-learning. Specifically, the optimization goal in the inner loop is to adapt the parameters in each candidate policy using the support data in sampled tasks, while the optimization objective in the outer loop alternates between learning the policy-specific parameters and learning the policy selection indicators.
During searching, we progressively decode the supernet from front to back stages, with fine-tune in between, based on a perturbation-based policy selection scheme~\cite{wang2021rethinking} that measures each policy's influence on the supernet.
At the end of the training, each network stage is associated with an adaptation policy with parameters learned.

To validate the effectiveness of our design, we conduct various experiments on multiple benchmark datasets, including the challenging cross-domain experiments. Our experiment results show that our searched-based model not only outperforms the random search baseline but also demonstrates significant performance advantages over many previous methods covered in our search space.
Our main contributions are summarized as follows:

\begin{itemize}
\itemsep 0cm 
	\item Our work is the first attempt to search meta-learning designs for few-shot learning tasks.
	\item We propose a hierarchical policy search space that covers many previous meta-learning algorithms.
	\item We develop a differentiable meta-learning policy searching algorithm that can conduct policy searching efficiently by meta-learning.
	\item Experiments on five popular datasets show that our method significantly outperforms the baselines and achieves new state-of-the-art results on many benchmarks.

\end{itemize}

Next we review some related work.

\section{Related work}
\textbf{Few-shot classification.}
Various few-shot learning paradigms were proposed in the literature \cite{simon2020adaptive, liu2020ensemble, zhang2020deepemd, hu20empirical, DhillonCRS20, ye2020few, yang2020dpgn, li2020adversarial, wang2020instance, wang2020trust, yu2020transmatch, liu2020negative
}.  
Metric-based approaches and optimization-based approaches are two dominating lines of efforts.
Metric-based methods~\cite{VinyalsBLKW16,SnellSZ17,SungCVPR2018,LiCVPR2019bestResult,ye2018learning,hou2019cross,dvornik2019diversity,prol2018cross,zhang2020deepemd,zhang2020deepemdv2} aim to learn a deep metric to inference data relations for predictions. Usually, once the model is learned, the parameters are fixed when it is deployed to inference in new tasks.
Therefore, metric based approaches have advantages in inference speed and often perform well in the extremely low-shot case. 
Optimization-based methods~\cite{FinnAL17,FinnNIPS2018,AntoniouICLR19,RaviICLR2017,LeeICML18,GrantICLR2018,ZhangNIPS2018MetaGAN,SunCVPR2019,LeeCVPR19svm,hu20empirical,lee2018gradient} aim to design effective learning paradigms for few-shot learning. For example, MAML~\cite{FinnAL17} aims to learn a good model initialization that can enable fast adaptations of network parameters in novel tasks. 
Chen~\etal~\cite{chen19closerfewshot} find that by simply pre-training the model weights with all training classes, many early works, such as ProtoNet~\cite{SnellSZ17}, MatchNet~\cite{VinyalsBLKW16}, and MAML~\cite{FinnAL17}  can be rejuvenated and reach state-of-the-art performance. In our experiments, we demonstrate the advantages of our model over these baselines that adopt a fixed policy across all network stages.
Besides image recognition, few-shot learning is also investigated in segmentation tasks~\cite{zhang2019canet,pgnet,liu2020crnet,chen2020compositional}.

\textbf{Network Architecture Search (NAS).}
Our work draws connections with the NAS literature \cite{liu2018darts,wang2021rethinking}, which aims for identifying effective building elements in the CNN structures. The most related work to ours is DARTS \cite{liu2018darts}, which applies continuous relaxation that transforms the discrete choice of architectures into architecture weights. 
In DARTS, different candidate operations together constitute a supernet that is optimized in a bi-level scheme, where a training set is used to learn the operation-specific parameters and a validation set is used to optimize the architecture weights. After training, the optimal operation is chosen by selecting the one with the largest architecture weights.
Despite its simplicity, many recent works question the effectiveness of DARTS \cite{Zela2020Understanding,wang2021rethinking,chen2020stabilizing,chen2021drnas}. For example, a simple random search baseline can outperform the architecture searched by DARTS \cite{liu2018darts}, and the searching favors parameter-free operations, \eg, skip connections \cite{Zela2020Understanding}. Recently, many improved designs are proposed to solve the issues in DARTS \cite{Zela2020Understanding,wang2021rethinking,chen2020stabilizing,chen2021drnas,li2020sgas,chu2020fair}.
For example, Wang \etal \cite{wang2021rethinking} find that the architecture weights may not be a good indicator for decoding the supernet, and propose a perturbation-based architecture selection approach. Specifically, after the training of supernet, the best operation is chosen based on how much each operation perturbs the supernet performance when it is removed. In our work, we also adopt such model discretization scheme to obtain the final policies. 

Recently, NAS for few-shot learning is explored in~\cite{elsken2020meta,doveh2019metadapt,lu2021neural,Lian2020Towards}, \emph{which aims to identify high-performing task-specific network structures in the few-shot learning tasks}. The difference of our work with them is that the searching goal in our framework is to find good parameter adaptation policies rather than architectures, and different candidate policy share the same architecture.

\begin{figure*}[t]
	\centering
	\includegraphics[width=1\linewidth]{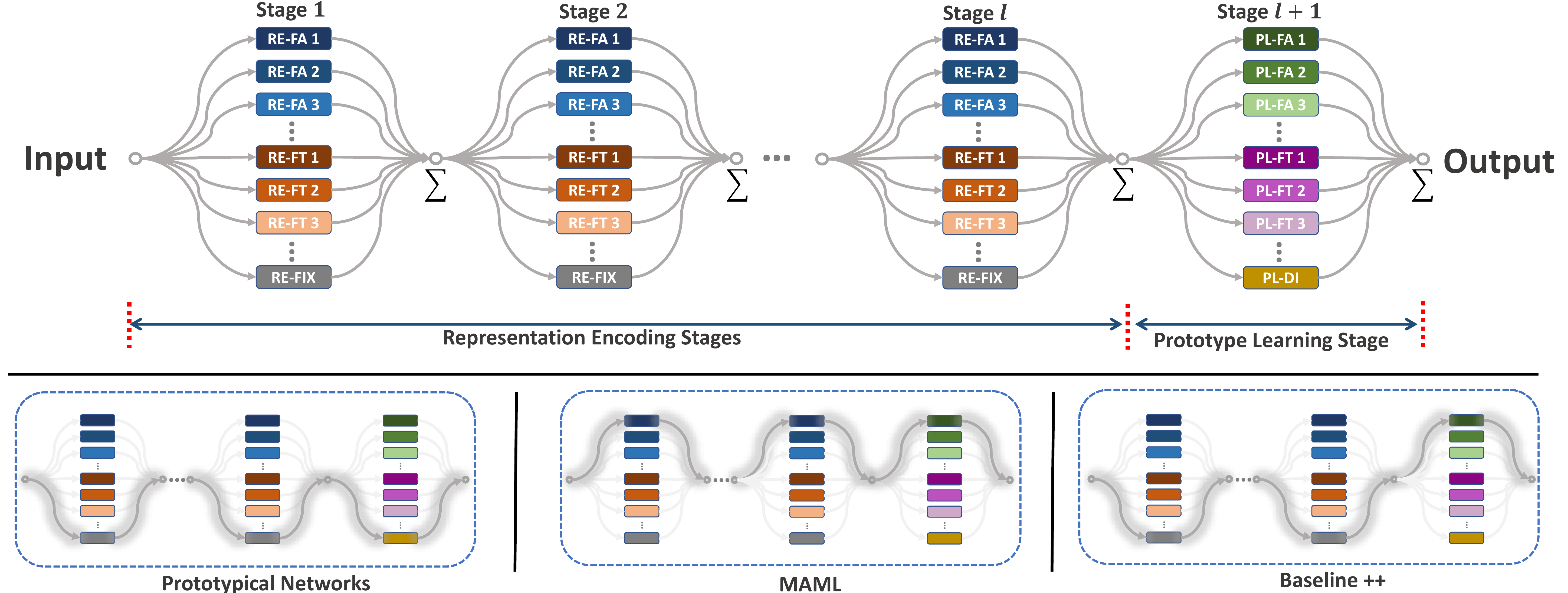}
	\caption{Our search-based framework for few-shot classification. We aim to search for a good adaptation policy at each network stage.	Our search space mainly include two parts: the policies for representation encoding and the policies for prototype learning. Please refer to Sec.~\ref{sec:PSS} for the explanation of different policies. Based on the continuous relaxation of different candidate policies at each stage, we construct a differentiable supernet that can be optimized end-to-end 
	(top). 
	Our search space covers many well-known few-shot learning algorithms, such as Prototypical Networks~\cite{SnellSZ17}, MAML~\cite{FinnAL17} and Baseline++~\cite{chen19closerfewshot} 
	(bottom). }
	\label{fig:main}

\end{figure*}

\section{Policy Search Space}\label{sec:PSS}
This section describes our meta-learning policy search space.
A standard CNN structure mainly include two components: the feature backbone that encodes the input image into representations and a classifier that classifies the data embedding. 
Therefore, we divide the search space in our framework into two parts: policies at the representation encoding (RE) stages (Sec.~\ref{sec:RE}) and the policies at the prototype learning (PL) stage (Sec.~\ref{sec:PL}).

\subsection{Search Space for Representation Encoding }
\label{sec:RE}

In a standard convolutional neural network, denoted by $\mathcal{G}(;)$, the backbone is used to encode input images into high-dimensional representations for classification by a sequence of convolutional layers. 
Due to the hierarchical design in CNNs, we can divide the sequential convolutional layers into several groups $\{g^1,g^2,... \}$, \eg,  4 layer blocks in ResNet. Our goal is to search an adaptation policy for each layer group in the context of few-shot classification tasks.
Specifically, given a layer group $g^l(;\theta)$, where $l$ is the index of the group and $\theta$ is the parameters in it, an adaptation policy is defined as the approach to adapt the parameters $\theta$ based on the support set $\mathcal{S}$ in a task $\mathcal{T}$, such that $\theta$ becomes $\hat{\theta}$. For each layer group in the backbone of a CNN, three kinds of candidate policies are involved: 

\textbf{I. Fixed Parameters (RE-FIX).} In RE-FIX, the parameters learned by training tasks are kept fixed without any adaptation in a novel task, \ie,  $\Hat{\theta} \leftarrow \theta$. Such design is widely seen in the metric-based approaches, where the learned data encoder on training tasks is directly reused to encode data in novel tasks. Similarly, baseline++~\cite{chen19closerfewshot} also freezes the pre-trained backbone in novel tasks and only fine-tunes the classifier. 

\textbf{II. Fine-Tuning the Weights (RE-FT).} In this case, the learned parameters of RE-FIX in the group can be fine-tuned with the support set $\mathcal{S}$ in novel tasks by stochastic gradient descent, \ie,
\begin{equation}
\label{eq:sgd}
 \Hat{\theta} = \theta - \beta \nabla_{\theta}\mathcal{L}_{\mathcal{T}}(\theta), 
\end{equation}
where $\beta$ is the learning rate and $\mathcal{L}$ is the loss function.
 By varying the hyper-parameters during fine-tuning, \eg, the learning rate or the number of iterations, we can obtain a collection of sub-candidates, such as a strong RE-FT policy with a large learning rate or a weak RE-FT policy that slightly fine-tunes the parameters with a small learning rate.

\textbf{III. Fast Adaptation (RE-FA).} Different from RE-FT that fine-tunes from the weights of RE-FIX, RE-FA fine-tunes the model from the model initialization that is meta-learned, as done in MAML \cite{FinnAL17}. In other words, the fine-tuning behavior in RE-FA influences the learning of model initialization, while RE-FT does not.
Similar to RE-FT, we can further obtain sub-candidates of RE-FA based on the adaptation hyper-parameters.

We search for an adaptation policy for each layer group $\mathcal{G}_l$ in the backbone. Therefore, dividing the backbone into $M$ stages results in a search space with $ [ 1+S_{\rm RE-FT}+S_{\rm RE-FA} ] ^M$ candidates for representation encoding, where $S$ denotes the number of sub-candidates in the policies.

\subsection{Search Space for Prototype Learning }
\label{sec:PL}
After encoding the input image into a vector representation $\mathbf{v} \in \mathbb{R}^{C}$ with the backbone, the classifier linearly projects the representation into the scores of each class $s_i$ with a weight matrix $\mathbf{W}\in \mathbb{R}^{N \times C}$, where $N$ is the number of classes and $C$ is the feature dimension.
From a prototype view of such operations, the weight matrix $\mathbf{W}$ essentially stores a collection of prototype vectors $[\mathbf{w}_1,\mathbf{w}_2,...,\mathbf{w}_N]^T$ for all classes, where $\mathbf{w}_i \in \mathbb{R}^{C}$, and the class score $s_i$ of a specific class $i$ is computed by the inner product between the data representation $\mathbf{v}$ and the class prototype $\mathbf{w}_i$, \ie, $s_i=\mathbf{w}_i\mathbf{v}^T$.
The inner product operation can also be replaced by other similarity metrics, such as negative L2 distance~\cite{SnellSZ17}, cosine similarity~\cite{VinyalsBLKW16} and deep Earth Mover's Distance~\cite{zhang2020deepemd}.
Based on such interpretation, we summarize the following policies as searching candidates to obtain the class prototypes for computing class scores:

\textbf{I. Data Initialization (PL-DI).} Metric-based meta-learning algorithms, such as Prototypical Networks~\cite{SnellSZ17}, can be seen as directly parameterizing the class prototypes with the data embeddings of support images. Typically, in the 1-shot case where each class has one support image, the encoded support data are directly used as the prototypes, while in the $k$-shot case, the averaged data embedding in each class is set as the class prototype~\cite{SnellSZ17}. Therefore, our first candidate is directly parameterizing the classifier with data embeddings without any adaptation.

\textbf{II. Fine-Tuning from Data Embeddings (PL-FT).} In this case, the data initialized prototypes in PL-DI are only served as the starting point for fine-tuning. 

\textbf{III. Fine-Tuning by Fast Adaptation (PL-FA).} Similar to the RE-FA policy for representation learning, the PL-FA fine-tunes the classifier from a meta-learned initialization of prototypes.

Likewise, we can obtain the strong or weak sub-candidates for PL-FT and PL-FA. We use cosine similarity to compute the class cores for all policies.
After involving the search space for prototypes, the whole search space $\mathcal{O}$ include $ [ 1+S_{\rm RE-FT}+S_{\rm RE-FA} ] ^M \times [ 1+S_{\rm PL-FT}+S_{\rm PL-FA} ]$ candidates, and the searching goal is to find a high-performing combination of policy candidates in the whole search space.

Our search space covers many well-known meta-learning algorithms, as shown in Fig.~\ref{fig:main}. For example, if all parameters are kept frozen for representation encoding and the classifier is directly parameterized by data embeddings (RE-FIX + PL-DI), the model becomes metric-based methods, such as Prototypical Networks~\cite{SnellSZ17} and Matching Networks~\cite{VinyalsBLKW16}; if all stages choose fast adaptation (RE-FA + PL-FA), the model becomes MAML; if the parameters in the backbone are kept frozen and the classifier is fine-tuned by fast adaptation (RE-FIX + PL-FA), it is close to baseline++~\cite{chen19closerfewshot}.

\section{Method}
In this part, we present our method for searching the policies in the aforementioned search space. We begin by introducing a continuous relaxation of different candidate policies that involve all policies into a differentiable supernet (Sec.~\ref{sec:relax}).
Then, we discuss how to optimize the model for searching~(Sec.~\ref{sec:optim}), and how to progressively decode the supernet (Sec.~\ref{sec:decode}).

\subsection{Continuous Relaxation of Policies}
\label{sec:relax}
 In order to search over a pool of discrete choice of meta-learning policies $\mathcal{O}^l$ on a specific stage $l$, we reuse the idea of continuous relaxation of individual choices, proposed in DARTS~\cite{liu2018darts}. 
At the beginning, each candidate policy $o^l_i$ in the search space $\mathcal{O}^l$ is associated with a normalized policy selection weight $\alpha^l_i$ as well as a copy of parameters $\theta^l_i$ at stage $l$, where,
 \begin{align}
               \sum_{i \in |\mathcal{O}^l|}\alpha^l_i=1, \alpha^l_i > 0,
 \end{align}
implemented by softmax. 
To obtain a continuous relaxation of discrete policies, we take the weighted sum of the outputs generated by different policies. Specifically, given the output of the previous stage $O^{l-1}$, the output of the current stage is computed by:
\begin{equation}
\label{eq:relax}
    O^{l}=\sum_{i \in |\mathcal{O}^l|}\alpha^l_i g^l(O^{l-1},\Hat{\theta^l_i}).
\end{equation}
Therefore, all candidate policies in different stages of the network together construct a differentiable supernet~\cite{liu2018darts}. In particular, $\Hat{\theta^l_i}$ is the parameters after adaptation in individual policies, while the policy selection weight $\alpha$ specifies the contribution of different candidate policies, which amounts to the architectures weights in DARTS~\cite{liu2018darts}.

\subsection{Optimization}
\label{sec:optim}
After constructing the supernet of the model, the next goal is to learn the parameters $\theta$ in the individual policies, as well as the policy selection weight $\alpha$. Recall that DARTS~\cite{liu2018darts} alternatively optimizes the architecture weights and the operation parameters with two disjoint sets, which approximates a bi-level optimization process.
With similar formulation, the optimization in our framework also alternates between two meta-learning objectiveness:
\begin{enumerate}\itemsep 0cm
  \item Update the parameter $\theta$ in different candidate policies with $\nabla_{\theta}. \mathcal{L}_{\mathcal{T}_{\text{t}}}(\Hat{\theta},\alpha)$
  \item Update the policy weights $\alpha$ with $\nabla_{\alpha} \mathcal{L}_{\mathcal{T}_{\text{v}}}(\Hat{\theta},\alpha)$.
\end{enumerate}
Algorithm~\ref{alg:episode} presents the pipeline of the optimization process in the form of pseudo-codes.
In comparison with the bi-level optimization scheme for NAS, there are two main difference: 1) Different from DARTS where the weights are learned on a specific task, the optimization in our framework is based on meta-learning, which samples tasks from two disjoint task domains $p(\mathcal{T}_{\text{A}})$ and $p(\mathcal{T}_{\text{B}})$, based on the training set and the validation set, respectively.
2) Each of the optimization objective above has a nested optimization problem. Specifically, in the inner loop of both optimization objectives, the goal is to adapt the policy weights $\theta$ to obtain task specific policy weights $\hat{\theta}$, while the outer loop alternatively optimizes the parameters in different policies $\theta$ and the policy selection weight $\alpha$, with $\nabla_{\theta}. \mathcal{L}_{\mathcal{T}_{\text{A}}}(\Hat{\theta},\alpha)$ and $\nabla_{\alpha} \mathcal{L}_{\mathcal{T}_{\text{B}}}(\Hat{\theta},\alpha)$, respectively. 
Moreover, the way to obtain the gradient with respect to the policies weights, \ie, $\nabla_{\theta}. \mathcal{L}_{\mathcal{T}_{\text{A}}}(\Hat{\theta},\alpha)$, varies in different policies. For example, since no adaptation operation is applied in RE-FIX, a closed-form expression of the gradients with respect to the parameters in RE-FIX can be obtained, while RE-FA and PL-FA must differentiate through the nested optimization trajectory  that requires the computation of gradients of gradients~\cite{FinnAL17}.
Noted that as all sub-candidates of RE-FT and PL-FT fine-tune the parameters from the pre-trained weights (RE-FIX) or data embeddings, the parameters ${\theta}$ for fine-tuning in these policies are generated online in each task, and hence there is no learnable parameter of these policies in the outer loop of 
Step~1.

\subsection{Decoding Discrete Policies }
\label{sec:decode}
During the supernet training, we progressively decode the supernet such that only one candidate policy is left in each stage.
Following~\cite{wang2021rethinking}, we adopt a perturbation-based decoding strategy that the strength of each poly is defined as how much it contributes to the performance of the supernet, which is implemented by masking out the path of each policy and observing the performance drop. The policy that leads to the largest accuracy drop on the validation set after being masked out is considered as the optimal policy in this stage.
We decode each stage one-by-one from front to back layers. After the decoding of each stage, we fine-tune the supernet with Algorithm~\ref{alg:episode} for a few episodes to recover the accuracy drop caused by discretization.

It is important to notice that the direct discretization will change the behaviors of fine-tuning based policies. Concretely, before decoding of the supernet, an SGD step in a policy $i$ at layer $l$ is
\begin{equation}
\begin{split}
     \Hat{\theta^l_i} & = \theta^l_i - \beta \frac{\partial \mathcal{L}_{\mathcal{S}}}{\partial \theta^l_i}\\
     & = \theta^l_i - \beta \frac{\partial \mathcal{L}_{\mathcal{S}}}{\partial  O^{l}}  \frac{\partial  O^{l}}{\partial g^l}  \frac{\partial  g^l}{\partial \theta^l_i}\\
     & = \theta^l_i - \beta \frac{\partial \mathcal{L}_{\mathcal{S}}}{\partial  O^{l}} \alpha^l_i  \frac{\partial  g^l}{\partial \theta^l_i}
\end{split}
\end{equation}
where $g^l$ is the output of the policy and $O_l$ is the output at the stage. 
The issue here is that, since the continuous relaxation in Eq.~\eqref{eq:relax} is discretized after decoding, the partial derivatives $\frac{\partial O^{l}}{\partial g^l}$ changes from $\alpha^l_i$ to $1$.
Hence, given the same gradients $\frac{\partial \mathcal{L}_{\mathcal{S}}}{\partial O^{l}}$ back-propagated to this stage, the discretization scales the adaptation strength of the policy by $1/\alpha^l_i$. To resolve such discrepancy, we fuse the policy selection weight $\alpha$ into the learning rate after decoding, \ie, $\beta \leftarrow \beta \alpha^l_i$ and the adaptation step after decoding becomes
\begin{equation}
\begin{split}
     \Hat{\theta^l_i}  & = \theta^l_i - (\beta \alpha^l_i)  \frac{\partial \mathcal{L}_{\mathcal{S}}}{\partial  O^{l}}  \frac{\partial  O^{l}}{\partial g^l}  \frac{\partial  g^l}{\partial \theta^l_i}\\
     &=\theta^l_i -  (\beta \alpha^l_i)  \frac{\partial \mathcal{L}_{\mathcal{S}}}{\partial  O^{l}}    \frac{\partial  g^l}{\partial \theta^l_i}
\end{split}
\end{equation}
As a result, the actual adaptation steps in the policy before and after decoding are identical, and the initial assigned learning rate only serves as the upper bound during searching.

\begin{algorithm}[t]
\caption{Optimization of the supernet to search for good adaptation policies.}
\label{alg:episode}
\SetAlgoLined
\SetKwInput{KwData}{Input}
\SetKwInput{KwResult}{Output}
 \KwData{$p(\mathcal{T}_A)$, $p(\mathcal{T}_B)$: two disjoint task distributions}
 \KwData{ $\mathcal{G(;\Uptheta,\alpha )}$: a supernet with pre-trained initialized weights in each policy}

\While{not done}{
\textcolor{blue}{\# Step 1: Optimize parameters $\Uptheta$ in the policies} \\
Sample a task $\mathcal{T}_A=\{\mathcal{S},\mathcal{Q}\}$ from  $p(\mathcal{T}_A)$\;
$\Hat{\Uptheta} \leftarrow \mathcal{G}(\mathcal{S}; \Uptheta,\alpha)$ , Adapt weights in each policy with the support set $\mathcal{S}$ \textcolor{blue}{(\# inner loop)}\;
Make predictions for the query set $\mathcal{Q}$ with  $\mathcal{G}(\mathcal{S}; \hat{\Uptheta},\alpha)$\;
Calculate loss  \textcolor{red}{$\nabla_{\Uptheta}. \mathcal{L}_{\mathcal{T}_{A}}(\Hat{\Uptheta},\alpha)$} and optimize \textcolor{red}{$\Uptheta$}\;

\textcolor{blue}{\# Step 2: Optimize policy selection weights $\alpha$} \\
Sample a task $\mathcal{T}_B=\{\mathcal{S},\mathcal{Q}\}$ from $p(\mathcal{T}_B)$\;
$\Hat{\Uptheta} \leftarrow \mathcal{G}(\mathcal{S}; \Uptheta,\alpha)$ , Adapt weights in each policy with the support set $\mathcal{S}$ \textcolor{blue}{(\# inner loop)} \;
Make predictions for the query set $\mathcal{Q}$ with $\mathcal{G}(\mathcal{S}; \hat{\Uptheta},\alpha)$\;
Calculate loss \textcolor{red}{$\nabla_{\alpha}. \mathcal{L}_{\mathcal{T}_{B}}(\Hat{\Uptheta},\alpha)$} and optimize \textcolor{red}{$\alpha$}\;
}

Decoding the supernet
\end{algorithm}

\section{Experiments}

\subsection{Dataset Statistics}
To validate the effectiveness of our framework, we conduct experiments on five benchmark datasets, including miniImageNet, tieredImageNet, Fewshot-CIFAR100 (FC100), CIFAR-FewShot (CIFAR-FS) and Caltech-UCSD Birds-200-2011 (CUB).

\textbf{\emph{mini}ImageNet.} \emph{mini}ImageNet is the most popular few-shot classification dataset, proposed in~\cite{VinyalsBLKW16}.
The dataset is built upon the ImageNet dataset~\cite{Russakovsky2015}, and contains 100 classes with 600 images in each class. 
The numbers of classes for training, validation and testing, are 64 ,16 and 20, respectively.

\textbf{\emph{tiered}ImageNet.} \emph{tiered}ImageNet is also a few-shot classification dataset build upon ImageNet, which includes 608 classes.   The splits of training(20), validation(6) and testing(8) classes are set according to the super-classes to enlarge domain gaps  between training and testing time.

\textbf{Fewshot-CIFAR100.} FC100 is a few-shot classification dataset build on CIFAR100~\cite{CIFAR100}.
Following the split division in~\cite{OreshkinNIPS18}, the training ,validation and testing sets include 60, 20, and 20 classes respectively.

\textbf{CIFAR-FewShot.} 
CIFAR-FS is also a few-shot classification dataset built upon CIFAR100, proposed in~\cite{DhillonCRS20}. It contains 64, 16, 20 classes for training, validation and testing, respectively.

\textbf{Caltech-UCSD Birds-200-2011.} CUB is a fine-grained bird classification dataset. Following~\cite{chen19closerfewshot}, we divide 200 classes into 100, 50 and 50 for training, validation and testing, respectively.

\subsection{Implementation Details}
We employ ResNet-12 as our network backbone to conduct all experiments. As there are four layer blocks in the ResNet backbone, we can naturally divide the network parameters into five stages, including four stages in the backbone and 1 stage in the classifier. 
We set two sub-candidates for RE-FT, RE-FA, PL-FT, PL-FA policies, including a strong version that fine-tunes parameters with the learning rate of 0.1 and a weak version with the learning rate of 0.01, and all the sub-candidates adapt the parameters for 10 epochs.
Therefore, our search space covers $(1+2+2) \times (1+2+2)^4=3125$ candidates totally.
Before the  optimization of the supernet, we pre-train a backbone with all data in the training set, and use the pre-trained weights  to initialize all policies at the representation learning stage.
We train the supernet with Alg.~\ref{alg:episode} for 1000 episodes and then start decoding from front to back layers. After decoding each stage, we fine-tune the supernet for 100 episodes to recover the accuracy drop caused by discretization. 
After all stages are decoded, we further fine-tune the networks for 2000 episodes.
Random scale, random crop and random horizontal flip are employed for data augmentation at training time.
All models in our experiment are evaluated with 600 testing episodes, and we report the average accuracy.

\begin{table}[t]
	\centering
\small
	\begin{tabular}{rccc}

		\Xhline{1.2pt}
		\multicolumn{1}{c}{Model}  &\textbf{ $1$-shot} & \textbf{ $5$-shot}  & \textbf{ $10$-shot} \\ \hline\hline
		ProtoNets~\cite{SnellSZ17}  & 57.89 & 78.75 & 82.66                  \\ 
		MatchNets~\cite{VinyalsBLKW16}  &  61.47 & 75.41 & 78.87   \\ 
		MAML~\cite{FinnAL17} &  59.58 & 75.80 & 78.20  \\ 
		Baseline++~\cite{chen19closerfewshot}  &  61.50 & 79.47 & 83.63    \\ 
		finetune & 44.81 & 68.77 & 76.94   \\
		Random search  & 64.10  & 77.97 &     80.94           \\  \hline
		\textbf{Ours}  &\textbf{65.91}    & \textbf{82.66}& \textbf{85.46}           \\ 	
	\Xhline{1.2pt}
	\end{tabular}%
	\caption{Comparison with baseline models for 5-way few-shot classification on \emph{mini}ImageNet dataset.  The searched model for 5-shot tasks is re-used to undertake 10-shot tasks in this experiment. Our searched policy outperforms baselines consistently on various tasks. }
	\label{table:baseline}
\end{table}

\begin{table}[t]
	
	\centering
	\small
			\vspace{7pt}
			\centering
	\begin{tabular}{rccc}

		\Xhline{1.2pt}
		\multicolumn{1}{c}{Model}  &\textbf{ $1$-shot} & \textbf{ $5$-shot}  & \textbf{ $10$-shot} \\ \hline\hline
					ProtoNets~\cite{SnellSZ17} &    45.52 & 66.80 & 72.29 \\
					MatchNets~\cite{VinyalsBLKW16}  & 47.41 & 63.63 &	68.83 \\
					MAML~\cite{FinnAL17} &   41.29 & 58.10 &	 62.18 \\
					Baseline++~\cite{chen19closerfewshot} &  47.79 & 70.01 &	76.13\\
					finetune & 39.49 & 67.88 & 78.60\\
					Random Search & 49.22 & 67.47 & 71.64  \\
					\hline
					\textbf{Ours}  & \textbf{53.80} & \textbf{72.43} & \textbf{81.05}  \\

					\Xhline{1.2pt}
				
					\vspace{0.01cm}
				\end{tabular}

		\caption{Cross-domain experiments on CUB dataset. Models trained with \emph{mini}ImageNet dataset are evaluated on multiple 5-way task settings. Our model performs well across all task settings and demonstrates performance advantages over all baselines.
		}
		\label{table:cross}	
\end{table}

\definecolor{mygray}{gray}{0.4}

\let\oldtiny\tiny
\renewcommand{\tiny}[1]{{\!\footnotesize{\textcolor{mygray}{#1}}}}

\begin{table*}[t]
\centering
	\begin{minipage}{1.0\textwidth}
		\begin{minipage}[t]{\textwidth}
			\vspace{7pt}
			\centering
			\resizebox{\textwidth}{!}{
				\begin{tabular}{ r lcccccccccccccc}
    \toprule
     \multirow{2.5}{*}{\textbf{Method}} & \multirow{2.5}{*}{\textbf{Backbone}} &  \multicolumn{2}{c}{\textbf{\emph{mini}ImageNet}} && \multicolumn{2}{c}{\textbf{\emph{tiered}ImageNet}} && \multicolumn{2}{c}{\textbf{FC100}} && \multicolumn{2}{c}{\textbf{CIFAR-FS}}\\ 
     \cmidrule{3-4} \cmidrule{6-7} \cmidrule{9-10} \cmidrule{12-13}
     && \textbf{$1$-shot} & \textbf{$5$-shot} && \textbf{$1$-shot} & \textbf{$5$-shot} && \textbf{$1$-shot} & \textbf{$5$-shot} && \textbf{$1$-shot} & \textbf{$5$-shot} \\
    \midrule
     TADAM \cite{OreshkinNIPS18} & ResNet-12 & 58.50 \tiny{$\pm$ $0.30$} & 76.70 \tiny{$\pm$ $0.30$} && --  & -- && 40.1 \tiny{$\pm$ $0.4$} & 56.1 \tiny{$\pm$ $0.4$} && -- & -- \\
     MTL \cite{SunCVPR2019} & ResNet-12 & 61.2 \tiny{$\pm$ $1.8$}  & 75.5 \tiny{$\pm$ $0.8$}  && 65.6 \tiny{$\pm$ $1.8$}  & 80.8 \tiny{$\pm$ $0.8$} && 45.1 \tiny{$\pm$ $1.9$}  & 57.6 \tiny{$\pm$ $1.0$}  && -- & -- \\
 FEAT \cite{ye2018learning} & ResNet-25${}^{\ddag}$ & 62.96 \tiny{$\pm$ $0.20$}  & 78.49 \tiny{$\pm$ $0.15$}  && --  & -- && --  & --  && -- & -- \\
 LEO~\cite{RusuICLR2019} & WRN-28-10${}^{\ddag}$ & 61.76 \tiny{$\pm$ $0.08$} & 77.59 \tiny{$\pm$ $0.12$} && 66.33 \tiny{$\pm$ $0.05$}  & 81.44 \tiny{$\pm$ $0.09$} && -- & -- && -- & -- \\
          Dhillon~\etal~\cite{DhillonCRS20} & WRN-28-10${}^{\ddag}$ & 57.73 \tiny{$\pm$ $0.62$}  & 78.17 \tiny{$\pm$ $0.49$}  && 66.58 \tiny{$\pm$ $0.70$}  & 85.55 \tiny{$\pm$ $0.48$} && 38.25 \tiny{$\pm$ $0.52$} & 57.19 \tiny{$\pm$ $0.57$} && 68.72 \tiny{$\pm$ $0.67$}  & 86.11 \tiny{$\pm$ $0.47$} \\
   MetaOptNet \cite{LeeCVPR19svm} & ResNet-12 & 62.64 \tiny{$\pm$ $0.82$} & 78.63 \tiny{$\pm$ $0.46$} && 65.99 \tiny{$\pm$ $0.72$} & 81.56 \tiny{$\pm$ $0.53$} && 41.1 \tiny{$\pm$ $0.6$} & 55.5 \tiny{$\pm$ $0.6$} && 72.0 \tiny{$\pm$ $0.7$} & 84.2 \tiny{$\pm$ $0.5$} \\
        CAN \cite{hou2019cross} & ResNet-12 & 63.85 \tiny{$\pm$ $0.48$}  & 79.44 \tiny{$\pm$ $0.34$}  && 69.89 \tiny{$\pm$ $0.51$}  & 84.23 \tiny{$\pm$ $0.37$} && --  & --  && -- & -- \\
    CTM \cite{LiCVPR2019bestResult} & ResNet-18${}^{\ddag}$ & 64.12 \tiny{$\pm$ $0.82$}  & 80.51 \tiny{$\pm$ $0.13$}  && 68.41 \tiny{$\pm$ $0.39$}   & 84.28 \tiny{$\pm$ $1.73$}  && --  & --  && -- & -- \\
            DSN-MR~\cite{simon2020adaptive} & ResNet-12 & 64.60 \tiny{$\pm$ $0.72$} & 79.51 \tiny{$\pm$ $0.50$} && 67.39 \tiny{$\pm$ $0.82$}  & 82.85 \tiny{$\pm$ $0.56$} && --  &-- &&  75.6 \tiny{$\pm$ $0.9$} &  86.2 \tiny{$\pm$ $0.6$} \\
                    Tian~\etal~\cite{tian2020rethinking} & ResNet-12 & 64.82 \tiny{$\pm$ $0.60$} & 82.14 \tiny{$\pm$ $0.43$} && 71.52 \tiny{$\pm$ $0.69$}  & 86.03 \tiny{$\pm$ $0.49$} && 44.6 \tiny{$\pm$ $0.7$} & 60.9 \tiny{$\pm$ $0.6$} && 73.9 \tiny{$\pm$ $0.8$} & 86.9 \tiny{$\pm$ $0.5$} \\
             Kim~\etal~\cite{kim2020model} & ResNet-12 & 65.08 \tiny{$\pm$ $0.86$} & 82.70 \tiny{$\pm$ $0.54$} && --  & -- && 42.31 \tiny{$\pm$ $0.75$} & 57.56 \tiny{$\pm$ $0.78$} && 73.51 \tiny{$\pm$ $0.92$} & 85.49 \tiny{$\pm$ $0.68$} \\
    DeepEMD~\cite{zhang2020deepemd} & ResNet-12 & 65.91 \tiny{$\pm$ $0.82$} & 82.41 \tiny{$\pm$ $0.56$} && 71.16 \tiny{$\pm$ $0.87$}  & \textbf{86.03} \tiny{$\pm$ $0.58$} && \textbf{46.47} \tiny{$\pm$ $0.78$} & \textbf{63.22} \tiny{$\pm$ $0.71$} && -- & -- \\

     \midrule
          \textbf{Ours}  & ResNet-12 &\textbf{65.91} \tiny{$\pm$ $0.83$} & \textbf{82.66} \tiny{$\pm$ $0.55$} && \textbf{73.52} \tiny{$\pm$ $0.88$} & \textbf{85.34} \tiny{$\pm$ $0.62$} && 45.60 \tiny{$\pm$ $0.81$} & 59.93 \tiny{$\pm$ $0.76$}&& 74.01 \tiny{$\pm$ $0.96$} & \textbf{86.03} \tiny{$\pm$ $0.62$} \\
          \textbf{Ours + MC}  & ResNet-12 & \textbf{67.14} \tiny{$\pm$ $0.80$} & \textbf{83.82} \tiny{$\pm$ $0.51$} && \textbf{74.58} \tiny{$\pm$ $0.88$} &\textbf{86.73} \tiny{$\pm$ $0.61$} && \textbf{46.40} \tiny{$\pm$ $0.81$}&  61.33 \tiny{$\pm$ $0.71$}&& \textbf{74.63} \tiny{$\pm$ $0.91$} &  \textbf{86.45} \tiny{$\pm$ $0.59$}  \\
    \bottomrule
    \multicolumn{10}{l}{ ${}^{\ddag}$ Different backbone with ours.}
    \vspace{0.01cm}
    \end{tabular}
				
		}

		\end{minipage}
	\caption{  Comparison with the state-of-the-art 5-way few-shot classification results  on \emph{mini}ImageNet, \emph{tiered}ImageNet, FC100, and CIFAR-FS datasets. \textbf{MC} denotes multi-crop testing. Our approach outperforms state-of-the-art performance on multiple datasets.
	}
	\label{table:sota}	
	\end{minipage}
\end{table*}

\begin{figure*}[t]
	\centering
	\includegraphics[width=1\linewidth]{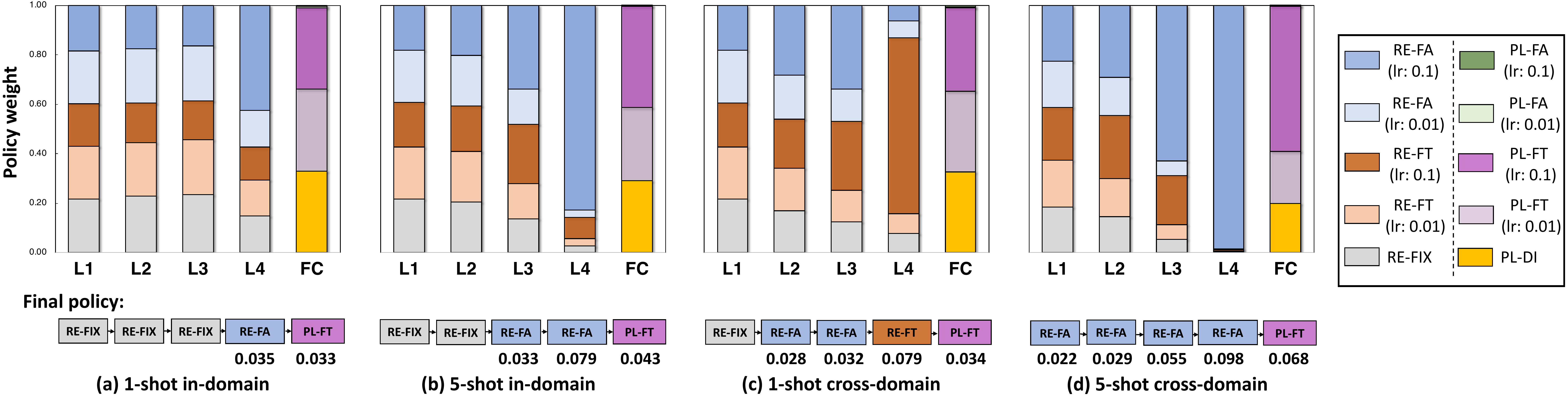}
	\caption{Visualization of the searched policies under different 5-way few-shot learning tasks. We plot the distributions of the policy selection weights before the initial decoding as well as the final policy at each network stage. Learning rates are noted below all fine-tune 
	based policies. Please refer to Sec.~\ref{sec:analysis} for our analysis.}
	\label{fig:visulization}
\end{figure*}

\subsection{Results and Analysis}
\label{sec:analysis}

\textbf{Visualization of searched models.}
We first present the visualization of the searched models under different few-shot learning tasks.
We plot the final selection of policies at each stage as well as the policy selection weights of different stages before the initial decoding in Fig.~\ref{fig:visulization}.
Noted that  the policy selection weights do not necessarily indicate the decoding selections, but we can observe the behavior of the searching model by comparing the selection distributions on different tasks or layers.
We have the following findings based on the visualizations:

\begin{enumerate}\itemsep 0cm
    \item 

What is shared across all tasks is that the prototypes that are fine-tuned with a relatively large learning rate based on data initialization (PL-FT) are always favored, while the prototypes that are fine-tuned from meta-learned initialization (PL-FA) are completely ignored.
This emphasizes the preference of the data initialization for prototypes learning in our searching model.
Meanwhile, the last representation encoding layer, \ie, layer 4, in all tasks also chooses adaptation with a relatively large learning rate (RE-FA or RE-FT).

\item 
%
We find that as the data-initialization-based policies  dominate the prototype learning stage in the supernet, the gradients propagated to the front layers are weak in the one-shot case. As a result, the outputs of different policies at the front layers, \eg, layer 1, are very similar, and the distribution is therefore close to uniform. Nevertheless, after the perturbation-based decoding, all  models except the one for the $k$-shot cross-domain task, choose to fix the parameters in the first  layer block.
%
\item 
By comparing the distributions and the final policies of different tasks, we can find that when the domain difference is large or the number of support data grows, more layers choose to be fine-tuned and the learning rate also increases. 
\end{enumerate}

\textbf{Comparison with baselines.}
In order to demonstrate the advantages of our design, we compare  our searched policies with the following baseline models that relate to our proposed search space:

\textbf{I. Prototypical Networks 
and 
Matching Networks}, the two representative metric-based methods in literature. Their difference is that Prototypical Networks~\cite{SnellSZ17} use negative L2 distance to compute class scores and the averaged embeddings as prototypes in the classifier, while Matching Networks~\cite{VinyalsBLKW16} use cosine similarity to compute class scores, and use individual data embeddings as prototypes and then fuse the scores generated by different prototypes from the same class.

\textbf{II. MAML~\cite{FinnAL17}}.  Based on our search space, all representation encoding stages adopt RE-FA policies and the prototype learning stage adopts the PL-FA policy. We choose the learning rates for the backbone and the classifier from $\{0.01, 0.1\}$ and report the optimal result. 

\textbf{III. Baseline++~\cite{chen19closerfewshot}}. Baseline++ freezes the pre-trained backbone and fine-tunes a cosine classifier. We fine-tune the classifier for 100 iterations for all experiments. The learning rates are chosen from $\{0.01, 0.1\}$, and we report the optimal result. 

\textbf{IV. Fine-tune.} We simply fine-tune the pre-trained model for 100 iterations with different learning rates for the backbone and the classifier. The learning rates are chosen from $\{0.01, 0.1\}$, and we report the optimal result. 

\textbf{V. Random Search.} We randomly sample the policies at each stage in the proposed search space to construct a model, and report the averaged performance of 10 sampled models. We omit the policy PL-FA (${\rm lr}=0.01$), as we find that 
it always generates poor results, no matter what policies are adopted at the prototype encoding stages. 

All the baseline models are pre-trained, and the results are presented in Table~\ref{table:baseline}.
As we can see, given the similar architectures shared by all compared models, our searched adaptation policy achieve optimal results on all task settings. 
In particular, our model outperforms the random search baseline consistently, which validates the effectiveness of our searching algorithm.

\textbf{Cross domain experiments.}
We next perform cross-domain experiments to further evaluate the effectiveness of our design, where the training data and testing data are sampled from different datasets. 
We evaluate the model trained with \emph{mini}Imagenet data on CUB dataset. 
Since CUB is a fine-grained classification dataset, there exists a domain gap between the training and testing tasks, which can better evaluate how well a few-shot learning algorithm adapt a model for novel tasks. In the cross-domain experiments, we use the validation set from the target domain as the set B in Alg.~\ref{alg:episode} to conduct searching.
As we can see from the results in Table~\ref{table:cross}, our model outperform baseline models consistently on different datasets with remarkable performance advantages. In particular, on the 10-shot tasks, we outperform the random search baseline by 9.41\% and baseline++ by 4.92\%.

\subsection{Comparison with State-of-the-Art Methods}
 To better position our method among the few-shot learning literature, we compare the results of our network with the state-of-the-art performance.
 We report the Top 1 accuracy with the 95\% confidence intervals on four benchmark datasets: \emph{mini}Imagenet, \emph{tiered}Imagenet, CIFAR-FS, and FC100.
As we find the predictions of the few-shot learners are often sensitive to the scales and shifts in input data, we also report the result that employs multi-crop testing, denoted by \textbf{MC}. Specifically, we simply re-use the data augmentation operations at training time  to scale and crop the query images  for 10 times and average the predicted logits of them as the final predictions. 
The result is shown in Table~\ref{table:sota}.
\emph{Although our method aims to solve a more general few-shot learning problems, we still obtain remarkable performance on the popular benchmarks.}
In particular, on popular \emph{tiered}Imagenet dataset, we obtain 74.58\% 1-shot accuracy, which outperforms previous 
state-of-the-art 
by 3.42\%.
 
\section{Conclusion}
In this paper, we have presented a search-based framework for few-shot learning classification that aims to find a good parameter adaptation policy at each network stage. 
 With a continuous relaxation of the discrete meta-learning policy, our searching model is differentiable and end-to-end trainable. We further develop a decoding algorithm that progressively select the optimal choice at each stage.
 Our designed search space covers many popular few-shot learning designs in literature.
 Extensive experiments validate the effectiveness of our design, and we obtain new state-of-the-art performance on multiple benchmarks.

 \section*{Acknowledgement}
 This research is supported by the National Research Foundation, Singapore under its AI Singapore Programme (AI\-SG Award No: AISG-RP-2018-003), and the MOE Ti\-er-1 Research Grants: RG28/18 (S), RG22/19 (S), and RG95/20.

{\small
\bibliographystyle{ieee_fullname}

\bibliography{final}
}

\end{document}